\documentclass{article}
\usepackage{ijcai20}

\usepackage{times}

\usepackage{soul}
\usepackage{url}
\usepackage[hidelinks]{hyperref}
\usepackage[utf8]{inputenc}
\usepackage[small]{caption}
\usepackage{graphicx}
\usepackage{amsmath}
\usepackage{booktabs}
\usepackage{amsthm}
\usepackage{amsmath,amssymb}
\usepackage{mathtools}
\usepackage{enumitem}

\DeclareMathOperator*{\argmax}{arg\,max}

\title{The Second Type of Uncertainty in Monte Carlo Tree Search}

\author{
Thomas M. Moerland$^{1,2}$\and
Joost Broekens$^{2}$\and
Aske Plaat$^{2}$\And
Catholijn M. Jonker$^{1,2}$\\
\affiliations
$^1$Interactive Intelligence, Delft University of Technology, The Netherlands \\
$^2$Leiden Institute of Advanced Computer Science, Leiden University, The Netherlands\\
}

\begin{document}

\maketitle

\begin{abstract}
Monte Carlo Tree Search (MCTS) efficiently balances exploration and exploitation in tree search based on count-derived uncertainty. However, these local visit counts ignore a second type of uncertainty induced by the size of the subtree below an action. We first show how, due to the lack of this second uncertainty type, MCTS may completely fail in well-known sparse exploration problems, known from the reinforcement learning community. We then introduce a new algorithm, which estimates the size of the subtree below an action, and leverages this information in the UCB formula to better direct exploration. Subsequently, we generalize these ideas by showing that loops, i.e., the repeated occurrence of (approximately) the same state in the same trace, are actually a special case of subtree depth variation. Testing on a variety of tasks shows that our algorithms increase sample efficiency, especially when the planning budget per timestep is small. 
\end{abstract}

\section{Introduction} \label{sec_introduction}
Monte Carlo Tree Search (MCTS) \cite{coulom2006efficient} is a state-of-the-art planning algorithm \cite{browne2012survey,chaslot2008monte}. The strength of MCTS is the use of statistical uncertainty to balance exploration versus exploitation \cite{munos2014bandits}. A popular MCTS selection rule is Upper Confidence Bounds for Trees (UCT) \cite{kocsis2006bandit,cazenave2007parallelization}, which explores based on the Upper Confidence Bound (UCB) \cite{auer2002finite} of the mean action value estimate. More recently, MCTS was also popularized in an iterated planning and learning scheme, where a low-budget planning iteration is nested in a learning loop. This approach achieved super-human performance in the games Go, Chess and Shogi \cite{silver2016mastering,silver2017mastering}.

However, the UCB selection rule only uses a local statistical uncertainty estimate derived from the number of visits to an action node. Thereby, it does not take into account how large the subtree below a specific action is. When we sample a single trace from a very large remaining subtree, we have much more remaining uncertainty than when we sample a trace from a very shallow subtree. However, standard MCTS cannot discriminate these settings. It turns out that MCTS can perform arbitrarily bad when the variation in subtree size between arms is large. 

We propose a solution to this problem through an extra back-up of {\it an estimate of the size of the subtree below an action}. This information is then integrated in an adapted UCB formula to better inform the exploration decision. Next, we show that {\it loops}, where the same state re-appears in a trace, can be seen as a special case of our framework. Our final algorithm, MCTS-T+, vastly increases performance in environments with variation in subtree depth and/or many loops, while performing at least on par to standard MCTS on environments that have less of these characteristics. Our experiments indicate that the benefits are mostly present 1) for single-player RL tasks with more early termination and loops, and 2) for lower computational budgets, which is especially relevant in real-time search with time limitations (e.g., robotics), and in iterated search and learning paradigms with small nested searches, e.g., in AlphaGo Zero \cite{silver2017mastering}.

The remainder of this paper is organized as follows. Section \ref{sec_preliminaries} provides essential preliminaries. Section \ref{sec_asymmetric} illustrates the problems caused by variation in subtree depth, and introduces a solution based on subtree depth estimation. Section \ref{sec_loops} identifies the problem of loops, and extends the algorithm of the previous section to MCTS-T+, which naturally deals with loops. The remaining sections \ref{sec_experiments}, \ref{sec_related}, \ref{sec_discussion} and \ref{sec_conclusion} present experiments, related work, discussion and conclusion, respectively. Code to replicate experiments is available from \emph{\url{https://github.com/tmoer/mcts-t.git}}.

\section{Preliminaries} \label{sec_preliminaries}

\subsection{Markov Decision Process}
We adopt a Markov Decision Process (MDP) \cite{sutton2018reinforcement} defined by the tuple $\{\mathcal{S},\mathcal{A},\mathcal{T},\mathcal{R},\gamma\}$. Here, $\mathcal{S} \subseteq \mathbb{R}^{N_s}$ is a state set, and $\mathcal{A} = \{1,2,..,N_a \}$ is a discrete action set. We assume that the MDP is deterministic, with transition function $\mathcal{T}: \mathcal{S} \times \mathcal{A} \to \mathcal{S}$ and reward function, $R: \mathcal{S} \times \mathcal{A} \to \mathbb{R}$. Finally, $\gamma \in (0,1]$ denotes a discount parameter. 

At every time-step $t$ we observe a state $s^t \in \mathcal{S}$ and pick an action $a^t \in \mathcal{A}$, after which the environment returns a reward $r^t = \mathcal{R}(s^t,a^t)$ and next state $s^{t+1} = \mathcal{T}(s^t,a^t)$. We act in the MDP according to a stochastic policy $\pi: \mathcal{S} \to P(\mathcal{A})$. Define the (policy-dependent)  state value $V^\pi(s_t) = \mathrm{E}_\pi[ \sum_{k=0}^\infty (\gamma)^k \cdot r^{t+k}]$ and state-action value $Q^\pi(s_t,a_t) = \mathrm{E}_\pi[ \sum_{k=0}^\infty (\gamma)^k \cdot r^{t+k}]$, respectively. Our goal is to find a policy $\pi$ that maximizes this cumulative, discounted sum of rewards.

\subsection{Monte Carlo Tree Search}
One approach to solving the MDP optimization problem from some state $s^t$ is through Monte Carlo Tree Search \cite{browne2012survey}. We here chose to illustrate our work with a variant of the PUCT algorithm \cite{rosin2011multi}, as used in AlphaGo Zero \cite{silver2017mastering}, but our methodology is equally applicable to other MCTS (select step) variants.
 
The tree consists of state nodes connected by action links. Each action link stores statistics $\{n(s,a),W(s,a),Q(s,a)\}$, where $n(s,a)$ is the visit count, $W(s,a)$ the cumulative return over all roll-outs through $(s,a)$, and $Q(s,a) = W(s,a)/n(s,a)$ is the mean action value estimate. MCTS repeatedly performs four subroutines \cite{browne2012survey}:

\begin{enumerate}[leftmargin=0.5cm]
\item {\bf Select}
We first descend the known part of the tree based on the tree policy rule:

\begin{equation}
\pi_\text{tree}(\cdot|s) = \argmax_a \Bigg[ Q(s,a) + c \cdot \frac{\sqrt{n(s)}}{n(s,a)} \Bigg] \label{eq_tree_policy}
\end{equation}

where $n(s)$ is the total number of visits to state $s$, and $c \in \mathbb{R}^+$ is a constant that scales exploration. The tree policy naturally balances exploration versus exploitation, as it initially prefers all actions (due to low visit count), but asymptotically only selects the optimal action(s). 

\item {\bf Expand} Once we encounter a child action edge which has not been tried before ($n(s,a)=0$), we expand the tree with a new leaf state $s_L$ according to the transition function. Subsequently, we initialize all the child links (actions) of the new leaf $s_L$. 

\item {\bf Roll-out} To obtain a fast estimate of the value of $s_L$, we then make a roll-out up to depth $D$ with some roll-out policy $\pi_\text{roll}$, for example a random policy, and estimate $\mathrm{R}(s_L) = \sum_{i=L}^{D} r(s_i,a_i)$. Instead of the roll-out, planning-learning integrations typically plug in a value estimate obtained from a learned value function \cite{silver2016mastering,silver2017mastering}.

\item {\bf Back-up} In the last step, we recursively back-up our value estimates in the tree. We recursively iterate through the trace, and update, for $i \in \{L, L-1, ..., 1, 0 \}$: 

\begin{equation}
\mathrm{R}(s_i,a_i) = r(s_i,a_i) + \gamma \mathrm{R}(s_{i+1},a_{i+1}). \label{eq_value_onpolicy}
\end{equation} 

where $\mathrm{R}(s_L,a_L) := \mathrm{R}(s_L)$, and subsequently set

\begin{align}
W(s_i,a_i) &\gets W(s_i,a_i) + R(s_i,a_i) \\
n(s_i,a_i) &\gets n(s_i,a_i) + 1 \\
Q(s_i,a_i) &\gets W(s_i,a_i)/n(s_i,a_i). \label{eq_mean}
\end{align}

\end{enumerate}

This procedure is repeated until the overall MCTS trace budget $N_{\text{trace}}$ is reached. We then recommend an action at the root, typically the one with the highest visitation count.

\begin{figure}[t]
  \centering
      \includegraphics[width=0.45\textwidth]{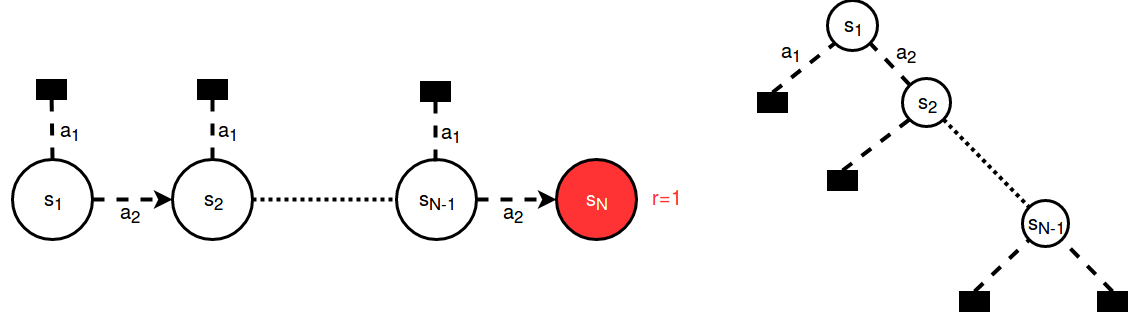}
  \caption{{\bf Left:} Chain domain. At each state we have two available actions: one action terminates the episode with reward 0, the other moves one step ahead in the chain with reward 0. Only the final state terminates the episode with reward 1. {\bf Right:} Search tree of the Chain domain.}
    \label{fig_asymmetric_chain}
\end{figure}

\begin{figure*}[ht]
  \centering
      \includegraphics[width=0.90\textwidth]{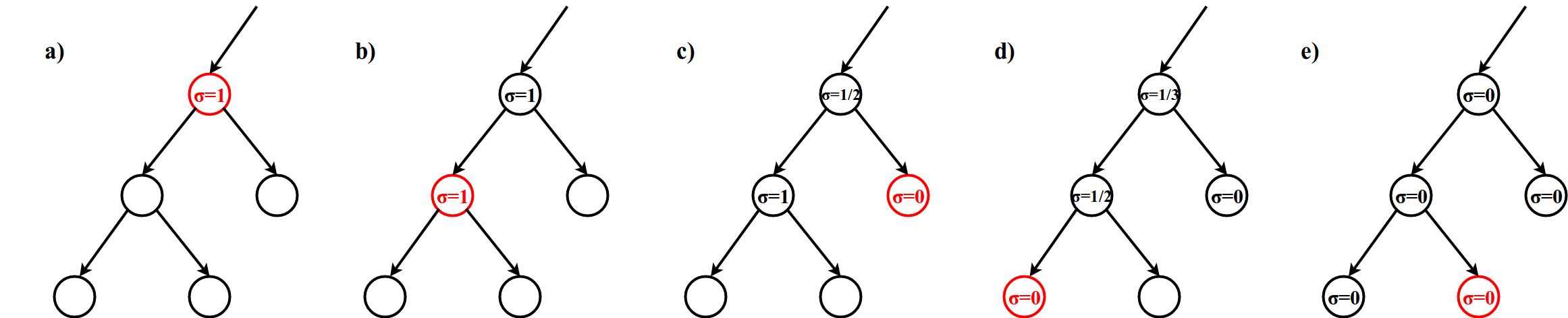}
  \caption{Process of $\sigma_\tau$ back-ups. Graphs a-e display subsequent estimates and back-ups of $\sigma_\tau$. In a) and b) we arrive at a non-terminal leaf node, of which the $\sigma_\tau$ automatically becomes 1. In the next subtree visit (c), we encounter a terminal leaf, and the uncertainty about the subtree at the subtree root decreases to $\frac{1}{2}$. In d) we encounter another terminal leaf. Because the back-ups are on-policy, we now estimate the root uncertainty as $\sigma_\tau = \frac{ (2 \cdot \frac{1}{2}) + (1 \cdot 0)}{2+1} = \frac{1}{3}$ (Eq. \ref{eq_sigma_backup}). Finally, at e) we enumerated the entire sub-tree, and the tree structure uncertainty at the subtree root is reduced to 0.}
    \label{fig_sigma_tree_backup}
\end{figure*}

\section{Variation in Subtree Size} \label{sec_asymmetric}
We now focus on a specific aspect that the above MCTS formuation does not account for: variation in the size of the subtree below actions (in the select step). Imagine we have two available actions in a certain state. The first action directly terminates the domain, and sampling it once therefore provides much information. The second action has a large subtree below it, and sampling it once only explores a single realization of all possible traces, with much more remaining uncertainty about the true optimal value. Now the key issue is: standard MCTS does not discriminate both cases, since it only tracks how often a node is visited, but completely ignores the size of the subtree below that action. 

Variation in subtree size is widespread in many single-player RL tasks. Examples include grid worlds \cite{sutton2018reinforcement}, exploration/adventure games (e.g. Montezuma's Revenge \cite{bellemare2013arcade}), shooting games (where in some arms we die quickly) \cite{kempka2016vizdoom}, and robotics tasks (where the robot breaks or environment terminates if we exceed certain physical limitations) \cite{brockman2016openai}. In the experimental section we test on different versions of such problems. 

When the subtree size below actions varies, then we can vastly gain efficiency by incorporating information about their size. For conceptual illustration, we will first focus on the Chain domain (Figure \ref{fig_asymmetric_chain}, left) \cite{osband2016generalization}, a well-known task from RL exploration research. The Chain is a long, narrow path with sparse reward at the end, which gives a very asymmetric tree structure that extends much deeper in one direction (Figure \ref{fig_asymmetric_chain}, right). 

The total number of terminating traces in this domain is $N+1$ for a Chain of length $N$. Exhaustive search therefore solves the task with $O(N)$ time complexity. Surprisingly, MCTS actually has {\it exponential} time complexity, $O(2^N)$, on this task. The problem is that MCTS receives returns of 0 for both actions at the root (since the chance of sampling the full correct trace is very small, $\sim \frac{1}{2^N}$. Therefore, MCTS keeps spreading its traces at the root, and recursively the same spreading happens at deeper nodes, leading to the exponential complexity. What MCTS lacks is information about the depth of the subtree below an arm. We empirically illustrate this behaviour in Sec. \ref{sec_chainresults}.

\subsection{MCTS with Tree Uncertainty Back-up (MCTS-T)}
We now extend the MCTS algorithm to make a soft estimate of the size of the subtree below an action, which we represent as the remaining uncertainty $\sigma_\tau(s) \in [0,1]$. For each state in the tree, we will estimate and recursively back-up $\sigma_\tau$, where  $\sigma_\tau(s)=1$ indicates a completely unexplored subtree below $s$, while $\sigma_\tau(s)=0$ indicates a fully enumerated subtree. 

We first define the $\sigma_\tau(s_L)$ of a new leaf state $s_L$ as:

\begin{equation}
\sigma_\tau(s_L) = \begin{cases}
0 &, \text{if }s_L\text{ is terminal} \\
1 &, \text{otherwise.}
\end{cases} \label{eq_sigma_tau}
\end{equation}

We then recursively back-up $\sigma_\tau$ to previous states in the search tree, i.e., we update $\sigma_\tau(s_i)$ from the uncertainties of its successors $\sigma_\tau(s_{i+1})$. We could use a uniform policy for this back-up, but one of the strengths of MCTS is that it gradually starts to prefer (i.e., more strongly weigh) the outcomes of good arms. We therefore weigh the $\sigma_\tau$ back-ups by the empirical MCTS counts. Moreover, if an action has not been tried yet (and we therefore lack an estimate of $\sigma_\tau$), then we initialize the action as if tried once and with maximum uncertainty (the most conservative estimate). Defining 

\begin{equation}
m(s,a) = \begin{cases}
n(s,a) &, \text{if } n(s,a) \geq 1 \\
1 &, \text{otherwise,}	
\end{cases}
\end{equation}

\begin{equation}
\sigma^\star_\tau(s') = \begin{cases}
\sigma_\tau(s') &, \text{if } n(s,a) \geq 1 \\
1 &, \text{otherwise,}
\end{cases}
\end{equation}

then the weighted $\sigma_\tau$ backup is 

\begin{equation}
\sigma_\tau(s) = \frac{\sum_{a} m(s,a) \cdot \sigma^\star_\tau(s')}{\sum_{a} m(s,a)} \label{eq_sigma_backup}
\end{equation}

for $s' = \mathcal{T}(s,a)$ given by the deterministic environment dynamics. This back-up process is illustrated in Figure \ref{fig_sigma_tree_backup}.

\paragraph{Modified select step} Small $\sigma_\tau$ reduces our need to visit that subtree again for exploration, as we already (largely) know what will happen there. We therefore modify our tree policy at node $s$ to:

\begin{equation}
\pi_{tree}(s) = \argmax_a \Big [ Q(s,a) + c \cdot \sigma_\tau(s') \cdot \frac{\sqrt{n(s)}}{n(s,a)}  \Big ] \label{eq_ucb_sigma}
\end{equation}

for $s' = \mathcal{T}(s,a)$ the successor state of action $a$ in $s$. The introduction of $\sigma_\tau$ acts as a prior on the upper confidence bound, reducing exploration pressure on those arms of which we have (largely) enumerated the subtree. 

\paragraph{Value back-up} The normal MCTS back-up averages the returns of all traces that passed through a node. However, the $\sigma_\tau$ mechanism, introduced above, puts extra exploration pressure on actions with a larger subtree below. Now imagine such a deep subtree has poor return. Then, due to $\sigma_\tau$, we will still visit the action often, and this will make the state above the action look too poor. When we are overly optimistic on the forward pass, we do not want to commit to always backing up the value estimate of the explored action. 

To overcome this issue, we specify a different back-up mechanism, that essentially recovers the standard MCTS back-up. On the forward pass, we track a second set of counts, $\tilde{n}(s,a)$, which are incremented {\it as if we acted according to the standard MCTS formula} (without $\sigma_\tau$):

\begin{equation}
\tilde{n}(s,a) \gets \tilde{n}(s,a) + \textbf{I} \big[a = \argmax_b Q(s,b) + c \cdot \frac{\sqrt{n(s)}}{n(s,b)}\big],
\end{equation}

where $\textbf{I}[\cdot]$ denotes the indicator function. We act according to Eq. \ref{eq_ucb_sigma}, but on the backward pass use the $\tilde{n}(s,a)$ counts for the value back-up:

\begin{equation}
Q(s,a) = \frac{\sum_{a'} \tilde{n}(s',a') \cdot Q(s',a')}{\tilde{n}(s')}, \label{eq_value_backup}
\end{equation}

for $s' = \mathcal{T}(s,a)$. This reweighs the means of all child actions according to the visit count they would have received in standard MCTS, which is the same as the standard MCTS back-up. 

Finally, we do no longer want to recommend an action at the root based on the counts, so we instead recommend the action with the highest mean value at the root.

\begin{figure*}[ht]
  \centering
      \includegraphics[width=0.90\textwidth]{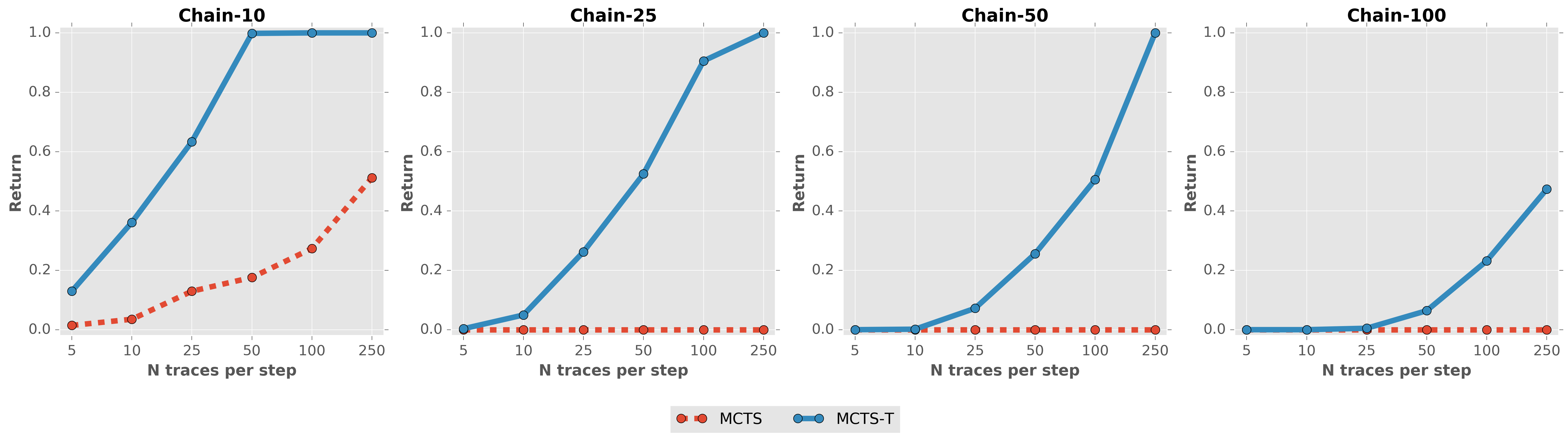}
  \caption{Comparison of vanilla MCTS (red) versus MCTS-T (blue) on the Chain domain of various lengths (progressing horizontally over the plots). Each plot displays computational budget per timestep (x-axis) versus average return per episode (y-axis). Results averaged over 25 episodes. We observe that MCTS-T achieves much higher returns in these domains with asymmetric termination and therefore variation in subtree depth.}
    \label{fig_results_chain}
\end{figure*}

\subsection{Results on Chain} \label{sec_chainresults}
Figure \ref{fig_results_chain} shows the performance of MCTS versus MCTS-T on the Chain (Fig. \ref{fig_asymmetric_chain}). Plots progress horizontally for longer lengths of the Chain, i.e., stronger asymmetry and therefore a stronger exploration challenge. In the short Chain of length 10 (Fig. \ref{fig_results_chain}, left), we see that both algorithms do learn, although MCTS-T is already more efficient. For the deeper chains of length 25, 50 and 100 (next three plots), we see that MCTS does not learn at all any more (flat red dotted lines), even for higher budgets. This illustrates the exponential sample complexity (in the length of the Chain) that MCTS starts to suffer from. In contrast, MCTS-T does consistently learn in the longer chains as well.

\section{Loops} \label{sec_loops}
We will next generalize the ideas about tree asymmetry to the presence of {\it loops} in a domain. A loop occurs when the same state appears twice {\it in the same trace} within a single search. In such cases, it never makes sense to further expand the tree below the second appearance. As an example, imagine we need to navigate three steps to the left. If we first plan one step right, then one step back left (a loop), then it does not make sense to continue planning to the left from that point. We could better plan to the left directly from the root itself.

There is an important conceptual difference between a loop and a transposition \cite{plaat1996exploiting}. Transpositions are ways of sharing information between states that were visited in other traces. In the above example, a transposition table stores the estimated value of going left in the start state. In contrast, a loop is a property within a single search, where information sharing has no benefit. Loops are especially frequent in single-player RL tasks, for example navigation tasks where we may step back and forth between two states. Note that the detection of loops does require full observability (since otherwise we do not know whether it we truly observe a repeated state, or something relevant changed in the background). 

\begin{figure}[t]
  \centering
      \includegraphics[width=0.45\textwidth]{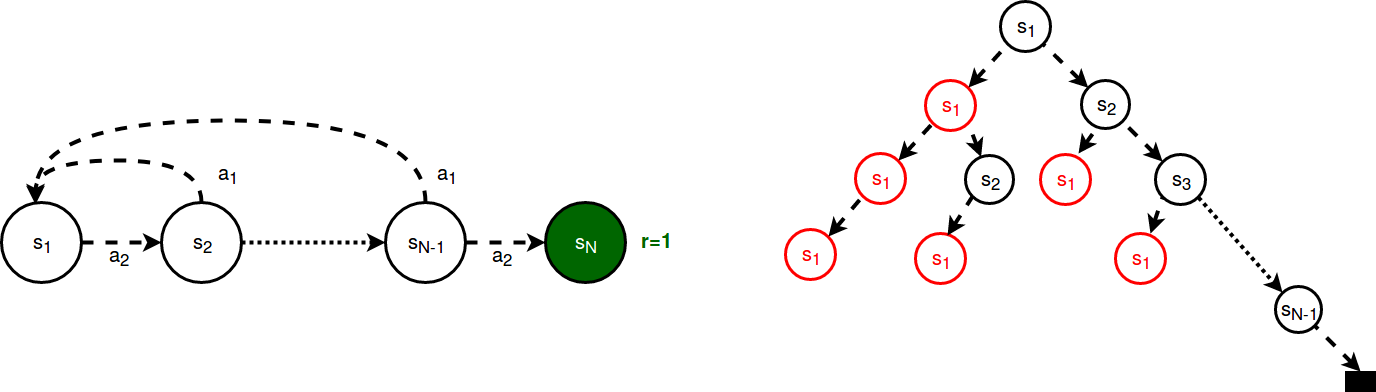}
  \caption{\small {\bf Left:} Chain domain with loops/cycles. {\bf Right:} Search tree of the cyclic Chain domain. Red nodes indicate a loop, i.e., the repetition of a state which already occurred in the trace above it.}
    \label{fig_cyclic_chain}
\end{figure}

We will illustrate the problem of loops with a variant of the Chain where the `wrong' action at each timestep returns the agent to state $s_1$ without episode termination (Figure \ref{fig_cyclic_chain}, left). When we now unfold the search tree (Figure \ref{fig_cyclic_chain}, right), we see that the tree is no longer asymmetric, but does have a lot of repeated appearances of state $s_1$. Standard MCTS cannot detect this problem, and will therefore repeatedly expand the tree in all directions.
 
\begin{figure*}[t]
  \centering
      \includegraphics[width=0.90\textwidth]{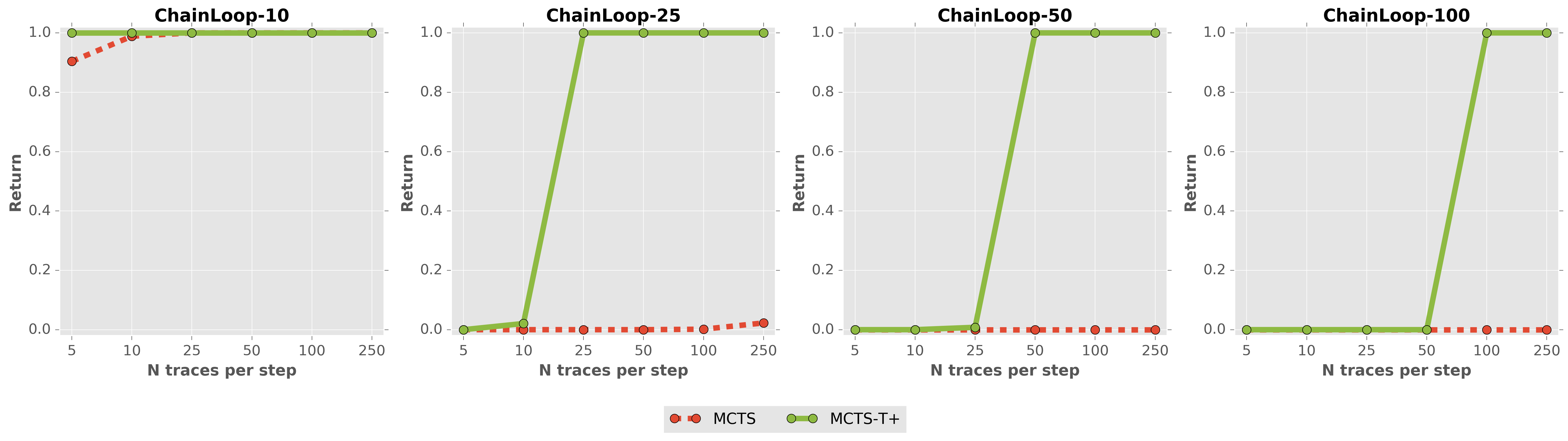}
  \caption{Comparison of MCTS (red) versus MCTS-T+ (green). MCTS-T+ uses tree uncertainty and loop blocking. Chain length progresses horizontally over the plots. Results averaged over 25 episodes. We observe that MCTS-T+ strongly outperform MCTS, which hardly incurs any reward on longer chains.}
    \label{fig_results_chain_loop}
\end{figure*}

\subsection{MCTS-T+: blocking loops.}
When we remove all the repeated visits of $s_1$, then we actually get the same tree as for the normal Chain again. This suggest that our $\sigma_\tau$ mechanism has a close relation to the appearance of loops as well. A natural solution is to detect duplicate states $s^\circ$ in a trace, and then set $\sigma_\tau(s^\circ)=0$. Thereby, we completely remove the exploration pressure from this arm, i.e., treat the looped state as if it has an empty subtree. 

The value/roll-out estimate of the duplicate state $\mathrm{R}(s^\circ)$ depends on the sum of reward in the loop $\mathrm{S}^\circ = \sum_{s,a \in g} r(s,a)$, where $g =\{s^\circ,..,s^\circ\}$ specifies the subset of the trace containing the loop. For infinite time-horizon problems with $\gamma=1$ (whose return is not guaranteed to be finite itself), we could theoretically repeat the loop forever, and therefore:

\begin{equation}
R(s^\circ) = \begin{cases}
\infty &, \text{if } \mathrm{S}^\circ \geq 0 \\
-\infty &, \text{if } \mathrm{S}^\circ \leq 0 \\
0 &, \text{if } \mathrm{S}^\circ = 0
\end{cases}
\end{equation}

For finite horizon problems, or problems with $\gamma<1$, we may approximate the value of the loop based on the number of remaining steps and the discount parameter. However, note that most frequently loops with a net positive or negative return are a domain artifact, as the solution of a (real-world) sequential decision making task is seldom to repeat the same action loop forever. 

In larger state spaces, exact loops are rare. We therefore check for approximate loops, where the looped state is very similar to a state above. We mark a new leaf state $s_L$ as looped when for any state $s_i$ above it, $L>i\geq0$, the L2-norm with the new expanded state is below a tunable threshold $\eta \in \mathbb{R}$:

\begin{equation} 
\| s_L - s_i \|_2 < \eta.
\end{equation}

Once a loop is detected, we set $\sigma_\tau=0$, and apply all methodology from the previous section. 

Note that a simpler solution to blocking loops could be to completely remove the parent action of a looped state from the tree. We present the above introduction to i) be robust against situations where the loop is relevant, and ii) to conceptually show what a loop implies: a state with an empty subtree below it ($\sigma_\tau=0$). 

\subsection{Results on Chain with loops}
We illustrate the performance of MCTS-T+ on the Chain with loops (Figure \ref{fig_cyclic_chain}). The results are shown in Figure \ref{fig_results_chain_loop}. We observe a similar pattern as in the previous section, where MCTS only (partially) solves the shorter chains, but does not solve the longer chains at all. In contrast, MCTS-T+ does efficiently solve the longer chains as well. Note that MCTS-T (without loop detection) does not solve this problem either (curves not shown), as the loops prevent any termination, and therefore all $\sigma_\tau$ estimates stay at 1.

\begin{figure*}[t]
  \centering
      \includegraphics[width=0.90\textwidth]{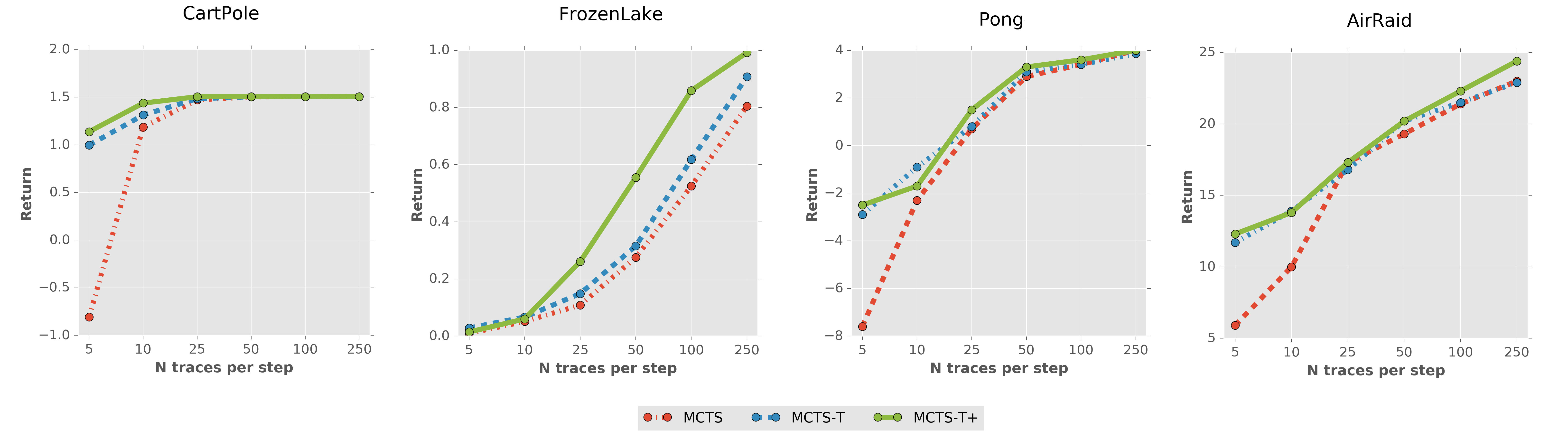}
  \caption{Learning curves on CartPole, FrozenLake, Pong and AirRaid. CartPole rewards are 0.005 for every timestep that the pole does not fall over, and -1 when the pole falls (episode terminates). We use the non-stochastic version of FrozenLake. The two Atari games, Pong and AirRaid, clip rewards to $[-1,1]$. All episodes last 400 steps, with a frameskip of 3 on the Atari games. There is no clear normalization criterion for the return scales on each domain, so we report their absolute values. Results averaged over 25 repetitions.}
    \label{fig_all_combined}
\end{figure*}

\section{Experiments} \label{sec_experiments}
The previous experiments, on the Chain and Chain with loops, present extreme cases of variation in subtree depth and the presence of loops. They are example cases to show the worst-case performance of MCTS in such scenarios, but are not very representative of most problems in the RL community. We therefore compare our algorithm to standard MCTS on several reinforcement learning tasks from the OpenAI Gym repository \cite{brockman2016openai}: CartPole, FrozenLake and the Atari games Pong and AirRaid. 

These results are visualized in Figure \ref{fig_all_combined}. We see that MCTS-T and MCTS-T+ consistently perform equal to or better than MCTS. This difference seems more pronounced for smaller MCTS search budgets. This seems to make sense, since the $\sigma_\tau$ machinery is especially applicable when we want to squeeze as much information out of our traces as possible. 

Note that the search budgets are relatively small compared to most tree search implementations. We will return to this point in the discussion. The computational overhead of MCTS-T itself is negligible (compared to the environment simulations). For MCTS-T+, loop detection does incur some cost in larger state spaces. In the worst case, on Atari, MCTS-T+ has $\sim$10\% increase in computation time.   
 
\section{Related Work} \label{sec_related}

The closest related work is probably MCTS-Solver \cite{winands2008monte}, designed for two-player, zero-sum games. In MCTS-Solver, once a subtree is enumerated, the action link above it is associated with its game-theoretical value $+\infty$ (forced win) or $-\infty$ (forced loss). It then uses specific back-up mechanisms (e.g., if one child action is a win, then the parent node is a win, and if all child actions are a loss, then the parent node is a loss). Compared to MCTS-Solver, our approach can be seen as a soft variant, where we gradually squeeze arms based on their estimated subtree size, instead of only squeezing completely once we fully enumerated the arm. Moreover, our approach is more generally applicable: it does not have any constraints on the reward functions (like win/loss), nor does it use back-up rules that are specific to two-player, zero-sum games. As such, MCTS-Solver would not be applicable to the problems studied in this paper. 

Other related work has focused on maintaining confidence bounds on the value of internal nodes, first introduced in B$^\star$ \cite{berliner1981b}. For example, score-bounded MCTS \cite{cazenave2010score} propagates explicit upper and lower bounds through the tree, and then prunes the tree based on alpha-beta style cuts \cite{knuth1975analysis}. This approach is only applicable to two-player games with minimax structure, while our approach is more general. Tesauro et al. \shortcite{tesauro2012bayesian} present a MCTS variant that propagates Bayesian uncertainty bounds. This approach is robust against variation in subtree size (not against loops), but requires priors on the confidence bounds, and will generally be quite conservative. One of the benefits of MCTS is that it gradually starts to ignore certain subtrees, without ever enumerating them, a property that is preserved in our approach.

While MCTS is a regret minimizing algorithm, a competing formulation, known as best-arm identification \cite{audibert2010best,kaufmann2017monte}, only cares about the final recommendation. Our approach also departs from the regret minimization objective, by putting additional exploration pressure on arms that have more remaining uncertainty. Finally, our solution also bears connections to RL exploration research papers that use the return distribution to enhance exploration \cite{moerland2018potential,tang2018exploration}, which implicitly may perform a similar mechanism as described in this paper. 

\section{Discussion} \label{sec_discussion}
This paper introduced MCTS-T+, an MCTS extension that is robust against variation in subtree size and loops. We will briefly cover some potential criticism and future extensions of our approach. 

From a games perspective, one could argue that our method is only useful in the endgame, when the search is relatively simple anyway (compared to the midgame). While this is true in two-player games, such as Go and Chess, many single-player reinforcement learning tasks, as studied in this paper, tend to have terminating arms right from the start (like dying in a shooting game), or many loops (like navigation tasks where we step back and forth). Our results are especially useful for the latter scenarios.  

Our methods seems predominantly beneficial with relatively small search budgets per timestep, compared to the budgets typically expended for search on two-player games. We do see three important ways in which our approach is relevant. First, real-time search with a limited time budget, as for example present in robotics applications, will benefit from maximum data efficiency. Second, we have recently seen a surge of success in iterated search and learning paradigms, like AlphaGo Zero \cite{silver2017mastering}, which nest a small search within a learning loop. Such approaches definitely require an effective small search. Finally, we believe our approach is also conceptually relevant in itself, since it identifies a second type type of uncertainty not frequently identified in MCTS, nor individually studied.

A limitation of our algorithm may occur when a sparse reward is hiding within an otherwise poorly returning subtree. In such scenarios, we risk squeezing out much exploration pressure based on initial traces that do not hit the sparse reward. However, MCTS itself suffers from the same problem, as its success also builds on the idea that the pay-offs of leafs in a subtree show correlation. Although MCTS does have asymptotic guarantees \cite{kocsis2006bandit}, it will generally also take very long on such sparse reward problems. This is almost inevitable, since these problems have such little structure that they technically require exhaustive search.

Note that in large domains without early termination, like the game of Go, MCTS-T will behave exactly like MCTS for a long time. As long as there is no expand step that reaches a terminal node, all $\sigma_\tau$ estimates remain at 1, and MCTS-T exactly reduces to MCTS. This gives the algorithm a sense of robustness: it exploits variation in subtree depth when possible, but otherwise automatically reduces to standard MCTS.

There are several directions for future work. First, the approach could be generalized to deal with stochastic and partially observable environments. Another direction would be to generalize information about $\sigma_\tau$, for example by training a neural network that predicts this quantity. Finally, the $\sigma_\tau$ mechanism may also suggest when a search can be stopped (e.g., all $\sigma_\tau \to 0$ at the root). Time management for MCTS has been studied before, for example by Huang et al. \shortcite{huang2010time}.

\section{Conclusion} \label{sec_conclusion}
This paper introduces MCTS-T+, an extension to vanilla MCTS that estimates the depth of subtrees below actions, uses these to better target exploration, and uses the same mechanism to deal with loops in the search. Empirical results indicate that MCTS-T+ performs on par or better than standard MCTS on several illustratory tasks and OpenAI Gym experiments, especially for smaller planning budgets. The method is simple to implement, has negligible computational overhead, and, in the absence of termination, stays equal to standard MCTS. It can be useful in single-player RL tasks with frequent termination and loops, real-time planning with limited time budgets, and iterated search and learning paradigms with small nested searches. Together, the paper also provides a conceptual introduction of a type of uncertainty that standard MCTS does not account for. 

\clearpage

\bibliographystyle{named}
\bibliography{example}

\begin{thebibliography}{}

\bibitem[\protect\citeauthoryear{Audibert and Bubeck}{2010}]{audibert2010best}
Jean-Yves Audibert and S{\'e}bastien Bubeck.
\newblock {Best arm identification in multi-armed bandits}.
\newblock 2010.

\bibitem[\protect\citeauthoryear{Auer \bgroup \em et al.\egroup
  }{2002}]{auer2002finite}
Peter Auer, Nicolo Cesa-Bianchi, and Paul Fischer.
\newblock {Finite-time analysis of the multiarmed bandit problem}.
\newblock {\em Machine learning}, 47(2-3):235--256, 2002.

\bibitem[\protect\citeauthoryear{Bellemare \bgroup \em et al.\egroup
  }{2013}]{bellemare2013arcade}
Marc~G Bellemare, Yavar Naddaf, Joel Veness, and Michael Bowling.
\newblock {The arcade learning environment: An evaluation platform for general
  agents}.
\newblock {\em Journal of Artificial Intelligence Research}, 47:253--279, 2013.

\bibitem[\protect\citeauthoryear{Berliner}{1981}]{berliner1981b}
Hans Berliner.
\newblock {The B* tree search algorithm: A best-first proof procedure}.
\newblock In {\em {Readings in Artificial Intelligence}}, pages 79--87.
  Elsevier, 1981.

\bibitem[\protect\citeauthoryear{Brockman \bgroup \em et al.\egroup
  }{2016}]{brockman2016openai}
Greg Brockman, Vicki Cheung, Ludwig Pettersson, Jonas Schneider, John Schulman,
  Jie Tang, and Wojciech Zaremba.
\newblock {Openai gym}.
\newblock {\em arXiv preprint arXiv:1606.01540}, 2016.

\bibitem[\protect\citeauthoryear{Browne \bgroup \em et al.\egroup
  }{2012}]{browne2012survey}
Cameron~B Browne, Edward Powley, Daniel Whitehouse, Simon~M Lucas, Peter~I
  Cowling, Philipp Rohlfshagen, Stephen Tavener, Diego Perez, Spyridon
  Samothrakis, and Simon Colton.
\newblock {A survey of monte carlo tree search methods}.
\newblock {\em IEEE Transactions on Computational Intelligence and AI in
  games}, 4(1):1--43, 2012.

\bibitem[\protect\citeauthoryear{Cazenave and
  Jouandeau}{2007}]{cazenave2007parallelization}
Tristan Cazenave and Nicolas Jouandeau.
\newblock {On the parallelization of UCT}.
\newblock In {\em {proceedings of the Computer Games Workshop}}, pages 93--101.
  Citeseer, 2007.

\bibitem[\protect\citeauthoryear{Cazenave and
  Saffidine}{2010}]{cazenave2010score}
Tristan Cazenave and Abdallah Saffidine.
\newblock {Score bounded Monte-Carlo tree search}.
\newblock In {\em {International Conference on Computers and Games}}, pages
  93--104. Springer, 2010.

\bibitem[\protect\citeauthoryear{Chaslot \bgroup \em et al.\egroup
  }{2008}]{chaslot2008monte}
Guillaume Chaslot, Sander Bakkes, Istvan Szita, and Pieter Spronck.
\newblock {Monte-Carlo Tree Search: A New Framework for Game AI.}
\newblock In {\em {AIIDE}}, 2008.

\bibitem[\protect\citeauthoryear{Coulom}{2006}]{coulom2006efficient}
R{\'e}mi Coulom.
\newblock {Efficient selectivity and backup operators in Monte-Carlo tree
  search}.
\newblock In {\em {International conference on computers and games}}, pages
  72--83. Springer, 2006.

\bibitem[\protect\citeauthoryear{Huang \bgroup \em et al.\egroup
  }{2010}]{huang2010time}
Shih-Chieh Huang, Remi Coulom, and Shun-Shii Lin.
\newblock {Time management for Monte-Carlo tree search applied to the game of
  Go}.
\newblock In {\em {2010 International Conference on Technologies and
  Applications of Artificial Intelligence}}, pages 462--466. IEEE, 2010.

\bibitem[\protect\citeauthoryear{Kaufmann and Koolen}{2017}]{kaufmann2017monte}
Emilie Kaufmann and Wouter~M Koolen.
\newblock {Monte-carlo tree search by best arm identification}.
\newblock In {\em {Advances in Neural Information Processing Systems}}, pages
  4897--4906, 2017.

\bibitem[\protect\citeauthoryear{Kempka \bgroup \em et al.\egroup
  }{2016}]{kempka2016vizdoom}
Micha{\l} Kempka, Marek Wydmuch, Grzegorz Runc, Jakub Toczek, and Wojciech
  Ja{\'s}kowski.
\newblock {Vizdoom: A doom-based ai research platform for visual reinforcement
  learning}.
\newblock In {\em {2016 IEEE Conference on Computational Intelligence and Games
  (CIG)}}, pages 1--8. IEEE, 2016.

\bibitem[\protect\citeauthoryear{Knuth and Moore}{1975}]{knuth1975analysis}
Donald~E Knuth and Ronald~W Moore.
\newblock {An analysis of alpha-beta pruning}.
\newblock {\em Artificial intelligence}, 6(4):293--326, 1975.

\bibitem[\protect\citeauthoryear{Kocsis and
  Szepesv{\'a}ri}{2006}]{kocsis2006bandit}
Levente Kocsis and Csaba Szepesv{\'a}ri.
\newblock {Bandit based monte-carlo planning}.
\newblock In {\em {ECML}}, volume~6, pages 282--293. Springer, 2006.

\bibitem[\protect\citeauthoryear{Moerland \bgroup \em et al.\egroup
  }{2018}]{moerland2018potential}
Thomas~M Moerland, Joost Broekens, and Catholijn~M Jonker.
\newblock {The Potential of the Return Distribution for Exploration in RL}.
\newblock {\em arXiv preprint arXiv:1806.04242}, 2018.

\bibitem[\protect\citeauthoryear{Munos and others}{2014}]{munos2014bandits}
R{\'e}mi Munos et~al.
\newblock {From bandits to Monte-Carlo Tree Search: The optimistic principle
  applied to optimization and planning}.
\newblock {\em Foundations and Trends{\textregistered} in Machine Learning},
  7(1):1--129, 2014.

\bibitem[\protect\citeauthoryear{Osband \bgroup \em et al.\egroup
  }{2016}]{osband2016generalization}
Ian Osband, Benjamin {Van Roy}, and Zheng Wen.
\newblock {Generalization and Exploration via Randomized Value Functions}.
\newblock In {\em {International Conference on Machine Learning}}, pages
  2377--2386, 2016.

\bibitem[\protect\citeauthoryear{Plaat \bgroup \em et al.\egroup
  }{1996}]{plaat1996exploiting}
Aske Plaat, Jonathan Schaeffer, Wim Pijls, and Arie {De Bruin}.
\newblock {Exploiting graph properties of game trees}.
\newblock In {\em {AAAI/IAAI, Vol. 1}}, pages 234--239, 1996.

\bibitem[\protect\citeauthoryear{Rosin}{2011}]{rosin2011multi}
Christopher~D Rosin.
\newblock {Multi-armed bandits with episode context}.
\newblock {\em Annals of Mathematics and Artificial Intelligence},
  61(3):203--230, 2011.

\bibitem[\protect\citeauthoryear{Silver \bgroup \em et al.\egroup
  }{2016}]{silver2016mastering}
David Silver, Aja Huang, Chris~J Maddison, Arthur Guez, Laurent Sifre, George
  {Van Den Driessche}, Julian Schrittwieser, Ioannis Antonoglou, Veda
  Panneershelvam, Marc Lanctot, et~al.
\newblock {Mastering the game of Go with deep neural networks and tree search}.
\newblock {\em Nature}, 529(7587):484--489, 2016.

\bibitem[\protect\citeauthoryear{Silver \bgroup \em et al.\egroup
  }{2017}]{silver2017mastering}
David Silver, Julian Schrittwieser, Karen Simonyan, Ioannis Antonoglou, Aja
  Huang, Arthur Guez, Thomas Hubert, Lucas Baker, Matthew Lai, Adrian Bolton,
  et~al.
\newblock {Mastering the game of go without human knowledge}.
\newblock {\em Nature}, 550(7676):354, 2017.

\bibitem[\protect\citeauthoryear{Sutton and
  Barto}{2018}]{sutton2018reinforcement}
Richard~S Sutton and Andrew~G Barto.
\newblock {\em {Reinforcement learning: An Introduction}}.
\newblock MIT press Cambridge, second edition, 2018.

\bibitem[\protect\citeauthoryear{Tang and Agrawal}{2018}]{tang2018exploration}
Yunhao Tang and Shipra Agrawal.
\newblock {Exploration by distributional reinforcement learning}.
\newblock {\em arXiv preprint arXiv:1805.01907}, 2018.

\bibitem[\protect\citeauthoryear{Tesauro \bgroup \em et al.\egroup
  }{2012}]{tesauro2012bayesian}
Gerald Tesauro, VT~Rajan, and Richard Segal.
\newblock {Bayesian inference in monte-carlo tree search}.
\newblock {\em arXiv preprint arXiv:1203.3519}, 2012.

\bibitem[\protect\citeauthoryear{Winands \bgroup \em et al.\egroup
  }{2008}]{winands2008monte}
Mark~HM Winands, Yngvi Bj{\"o}rnsson, and Jahn-Takeshi Saito.
\newblock {Monte-Carlo tree search solver}.
\newblock In {\em {International Conference on Computers and Games}}, pages
  25--36. Springer, 2008.

\end{thebibliography}

\end{document}